\documentclass[journal,twoside,web]{ieeecolor}
\usepackage{tmi}
\usepackage{cite}
\usepackage{amsmath,amssymb,amsfonts}
\usepackage{graphicx}
\usepackage{textcomp}
\usepackage{algorithm}
\usepackage{algorithmic}
\usepackage{hyperref}
\usepackage{amsmath}
\usepackage{color}
\usepackage{multirow}
\usepackage[T1]{fontenc}
\usepackage{soul}

\usepackage{ulem}
\usepackage{bbm}
\def\BibTeX{{\rm B\kern-.05em{\sc i\kern-.025em b}\kern-.08em
    T\kern-.1667em\lower.7ex\hbox{E}\kern-.125emX}}
\begin{document}
\title{Imbalanced Medical Image Segmentation with Pixel-dependent Noisy Labels}
\author{Erjian Guo, Zicheng Wang, Zhen Zhao, and Luping Zhou, \IEEEmembership{Senior Member, IEEE}
\thanks{The project is supported by Australian Research Council (ARC) DP200103223.}
\thanks{E.Guo, Z.Wang, Z.Zhao, and L.Zhou are with School of Electrical and Computer Engineering, University of Sydney, Australia(email: eguo9622@uni.sydney.edu.au; zwan4733@uni.sydney.edu.au; zhen.zhao@sydney.edu.au; luping.zhou@sydney.edu.au).}}
\maketitle

\begin{abstract}
Accurate medical image segmentation is often hindered by noisy labels in training data, due to the challenges of annotating medical images. Prior research works addressing noisy labels tend to make class-dependent assumptions, overlooking the pixel-dependent nature of most noisy labels. Furthermore, existing methods typically apply fixed thresholds to filter out noisy labels, risking the removal of minority classes and consequently degrading segmentation performance. To bridge these gaps, our proposed framework, Collaborative Learning with Curriculum Selection (CLCS), addresses pixel-dependent noisy labels with class imbalance. CLCS advances the existing works by i) treating noisy labels as pixel-dependent and addressing them through a collaborative learning framework, and ii) employing  a curriculum dynamic thresholding approach adapting to model learning progress to select clean data samples to mitigate the class imbalance issue, and iii) applying a noise balance loss to noisy data samples to improve data utilization instead of discarding them outright. Specifically, our CLCS contains two modules: Curriculum Noisy Label Sample Selection (CNS) and Noise Balance Loss (NBL). 
In the CNS module, we designed a two-branch network with discrepancy loss for collaborative learning so that different feature representations of the same instance could be extracted from distinct views and used to vote the class probabilities of pixels. Besides,  a curriculum dynamic threshold is adopted to select clean-label samples through probability voting. In the NBL module, instead of directly dropping the suspiciously noisy labels, we further adopt a robust loss to leverage such instances to boost the performance. We verify our CLCS on two benchmarks with different types of segmentation noise. Our method can obtain new state-of-the-art performance in different settings, yielding more than 3\% Dice and mIoU improvements. Our code is available at https://github.com/Erjian96/CLCS.git.

\end{abstract}

\begin{IEEEkeywords}
Medical image segmentation, Learning with label noise.
\end{IEEEkeywords}

\section{Introduction}
\label{sec:introduction}
\IEEEPARstart{I}{mage} segmentation is an important task in medical image analysis, with significant potential for various clinical applications. Deep learning algorithms based on convolutional neural networks (CNNs) have demonstrated remarkable advancements in medical image segmentation~\cite{zhang2020generalizing,tan2022retinal}. 
However, such impressive success is closely dependent on large amounts of clean training data with precise pixel-level annotations. 
In practical medical applications, obtaining high-quality pixel-level annotations is a challenging and labor-intensive task, due to the lack of experienced annotators, visual ambiguity in object boundaries, and limited budgets~\cite{guo2022joint,li2021superpixel}.
As a result, the medical training dataset inevitably contains noisy labels, and their presence may mislead the segmentation model to memorize wrong semantic correlations, reducing the generalizability of models. Hence, it is important to develop robust medical image segmentation techniques adept at handling noisily labeled data for model training. 

The issue of noisy labels has been extensively studied in image classification tasks. However, the impacts of pixel-wise label noise on segmentation tasks, particularly in the realm of medical image analysis, have not been thoroughly investigated. 
In contrast to the extensively explored class-dependent noise, pixel-wise label noise defies prior distribution assumptions and lacks fixed noise patterns. 
This type of noise is not only practical but also prevalent in real-world medical scenarios.
One apparent solution is to employ general regularization techniques to address noisy labels,
which helps in preventing the model from overfitting noisy patterns. These techniques include early stopping, drop out,  and label smoothing. 
Early stopping is commonly applied in noisy classification tasks,
which relates to the commonly known "memorization effect", where deep neural networks tend to first memorize and fit majority (clean) patterns and then overfit minority (noisy) patterns~\cite{arpit2017closer}. Early stopping  therefore can avoid over-fitting noisy labels. However, stopping the training prematurely can result in underfitting clean labels. Another technique to employ is spatial label smoothing regularization, often utilized for class-wise label refurbishments~\cite{lukasik2020does}. Label smoothing prevents the network from becoming overly confident, which, in turn, helps reduce the extent to which the model overfits noisy label data. While these general regularization methods are effective to some extent for classification tasks, they prove to be insufficient for segmentation tasks with pixel-level noise.

Recent dominant approaches to address noisy labels for medical image segmentation can be broadly categorized into two groups. \textbf{The first group} denoises class-wise label noise by using the noise transition matrix (NTM) and robust losses. The works~\cite{xia2019anchor,li2021provably} modeled class-wise noisy label distribution through either the confusion matrix or noise transition matrix (NTM) and then leveraged the modeled distribution for loss correction. However, estimating the noisy label distribution becomes increasingly challenging when the number of classes increases or the noise rate escalates~\cite{han2018co} in the segmentation tasks. More importantly, 
NTM assumes that the probability of a pixel belonging to a class being corrupted into one of the other classes is consistent for all the pixels in the dataset. This assumption is too restrictive to hold true in practical segmentation tasks, where varied and complex noise patterns are more common. Consequently, the pixel-wise noise may not conform to specific distribution assumptions and cannot be well addressed by NTM-based methods.
\textbf{The second group} denoises pixel-wise label noise using Co-Training frameworks. The works~\cite{han2018co,liang2022review} leveraged confidence scores or loss values to select reliable labels for co-training through two-branch networks. These methods usually define a fixed threshold to exclude pixels with low-confidence labels or with large losses. However, using fixed thresholds, these methods fail to either consider the evolving nature of the model during the training procedure or distinguish the difficulty of segmenting varied classes. As a result, these methods may exclude the pixels of small-sized objects that are inherently challenging to segment, making the model under-represent small-sized objects. Consequently, the issue of class imbalance, which is an inherent problem of medical image segmentation, will be exacerbated. 
Research in pixel-wise noise of class-imbalanced cases, especially in medical imaging segmentation tasks, has not been well explored.

To bridge these research gaps,  we propose a Collaborative Learning with Curriculum Selection (CLCS) framework for imbalanced medical image segmentation with pixel-wise noisy labels. Our approach employs a two-branch network that collaboratively distinguishes between clean and noisy samples. The core idea behind this collaborative framework is to harness the complementary nature of these two branches, enabling them to rectify each other's errors. Different sub-networks are encouraged to provide diverse perspectives on the same instances, resulting in varied predictions. 
To prevent the two sub-networks from converging to the same prediction during training, we introduce a discrepancy loss that compels them to make decisions from substantially distinct viewpoints. 
We can then vote on each pixel among these models' predictions as well as the original labels to select potentially clean labels.
To differentiate samples with clean and noisy labels, we design a dynamic threshold to select pixels with clean labels according to the model’s learning status and the sizes of classes, dubbed Curriculum Noisy Label Sample Selection (CNS). 
Similar to the idea of learning from easy to hard in curriculum learning, our dynamic threshold in CNS adapts as the model learns, commencing with easily distinguishable clean pixels and gradually incorporating more challenging noisy pixels. 
To alleviate the problem of imbalanced classes, with the progression of the model, our dynamic threshold is updated based on the proportion of pixels for each class according to the model's current predictions. 
Moreover, unlike conventional methods that simply discard noisy samples, we argue that relying on threshold-based techniques cannot perfectly distinguish the clean labels from the noisy ones, and consequently the detected noisy labels will inevitably contain genuinely clean labels. To address this issue and further improve the data utility, we adopt a Noise Balance Loss (NBL)  to incorporate the detected noisy samples to further improve segmentation.

Our contributions can be summarized as follows.
\begin{enumerate}
    \item We delve into medical image segmentation tasks marked by pixel-wise noisy labels while also tackling the intrinsic challenges of class imbalance exacerbated by the presence of noisy labels. This particular scenario is less explored but holds importance for real-world medical imaging applications. To address this problem, we propose a Collaborative Learning with Curriculum Selection (CLCS) framework to reduce the influence of noisy labels for medical image segmentation with imbalanced classes.
    \item We propose a Curriculum Noisy label sample Selection (CNS) module, which utilizes a two-branch network with a discrepancy loss for sample voting and adopts a dynamic threshold mechanism adaptive to the model learning progress and the imbalanced classes to differentiate clean and noisy labels.
    \item  Rather than directly discarding the detected noisy samples that could potentially contain genuinely clean ones such as those from the minority classes, our proposed Noise Balance Loss (NBL) module further applies a robust loss to noisy labels to incorporate them into the training, which not only elevates the efficiency of data utilization but also mitigates the class imbalance issue in a way different from the conventional solutions.
    \item  The proposed method demonstrates significant performance improvements on two real-world medical image datasets with different noise types consistently.
    \end{enumerate}
\section{Related work}
In this section, we briefly discuss learning with noisy labels on medical image segmentation tasks, the collaborative learning framework, curriculum learning strategy and noise type for segmentation tasks, which are related to our work.

\subsection{Learning with Noise Labels}
Learning with noise labels has been extensively studied in recent works~\cite{nishi2021augmentation,wang2018iterative}. Two common groups of approaches are employed to address the problem of noisy labels in segmentation tasks: denoising class-wise labels and denoising pixel-wise labels. For the first group, researchers employ various techniques such as regularization, noise transition matrix, and robust losses to mitigate the impact of noisy labels without explicitly discarding the noisy labeled data. These types of approaches are similar to those handling noisy labels for classification tasks. The objective is to effectively reduce the influence of label noise during training while preserving as much useful information as possible. For example, a regularization technique was proposed by Wei et al.~\cite{wei2020combating}. Some methods estimated noise transition matrices to correct label noise in a class-wise manner~\cite{hendrycks2018using}. Other popular methods reduced the effects of noisy labels by using noise-tolerant loss functions~\cite{chang2017active, zhang2018generalized}. However, these methods tend to perform poorly when confronted with high noise rates and a large number of classes in segmentation tasks. Moreover, estimating the distribution relationship between noisy and clean data based on categories is impractical in segmentation tasks. Alternative methods focused on selecting cleanly labeled samples and training models exclusively on these clean instances~\cite{ding2018semi,han2018co}. Notably, the selection criteria for clean samples differ across these methods. In our approach, we employ a noise-robust dynamic voting strategy to select clean label data according to model learning status and employ a robust loss function to extract useful information from the remaining data rather than directly discarding them.

\subsection{Collaborative Learning}
Collaborative learning aims to learn two or more distinct and divergent feature extractors for the same instances, which has also been employed to handle noisy label tasks.
For example, Co-Teaching~\cite{chen2021semi}, a popular two-branch approach, involves two subnets providing each other with different and complementary information. It uses small loss tricks to select confident clean samples without considering the original labels. This two-branch framework has been used in segmentation tasks, but the two sub-nets are easy to collapse, i.e., converging to the same predictions. To prevent this issue, some methods~\cite{ouali2020semi} used feature-level perturbations to produce different inputs of the decoders, preventing the decoders from collapsing into each other. However, using artificial perturbations on encoders makes it easy to corrupt learning reasonable features from the encoders. The work~\cite{chen2021semi} imposed consistency on two segmentation networks perturbed with different initializations for the same input image to generate predictions of different views. However, this method only guarantees divergence at the beginning of training, while the predictions of the two-branch network gradually converge as the training progresses. In contrast, in our two-branch framework, we propose a discrepancy loss to prevent the collapse of two sub-nets. It compels the two sub-nets to learn from distinct views to vote for the label prediction, which achieves better performance. Moreover, our collaborative learning with confident voting combines the predictions from both the two branches and the original labels, leading to a stable and accurate prediction and reducing the influence of confirmation bias.
\begin{figure*}[t]
\begin{center}
\centerline{\includegraphics[width=1\textwidth]{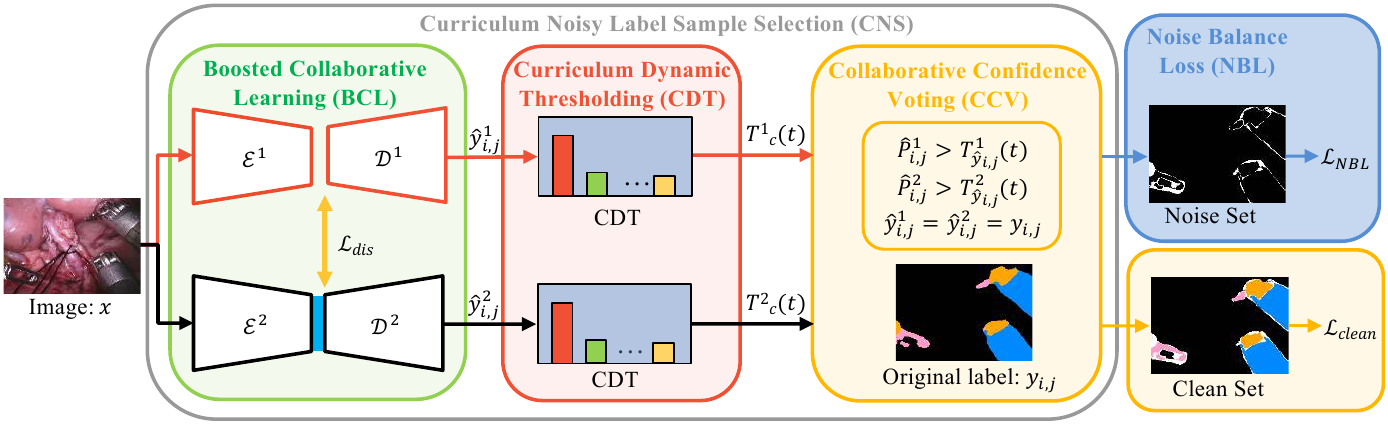}}
\end{center}
\vspace{-6mm}
\caption{\textbf{Overview of Collaborative Learning with Curriculum Selection (CLCS).} An input image is processed by each of the two network branches individually to generate predictions from distinct viewpoints, facilitated by a discrepancy loss. Leveraging the predictions from the two branches and the original label, the model groups the pixels into a clean set and a noise set by the Curriculum Noisy Label Sample Selection (CNS) module. The network predictions and original labels are integrated through a curriculum dynamic threshold with a robust voting strategy. The model is trained by minimizing the supervised cross-entropy loss for the clean set and the Noise Balance Loss (NBL) for the noise set. The blue block in the BCL module is the mapping layer.}
\label{fig2}
\end{figure*}


\subsection{Noise Type for Segmentation Tasks}
Most existing methods to deal with artificially synthesized noisy labels were initially proposed for image classification tasks~\cite{karimi2020deep}. In these classification tasks, synthesized noisy labels typically focus on both symmetric and asymmetric label noise. Symmetric label noise involves flipping real labels to other labels with equal probability, while asymmetric label noise entails flipping real labels to other labels based on fixed rules. However, when it comes to segmentation tasks, symmetric and asymmetric noise models appear unreasonable. Unlike the traditionally synthesized label noise for classification, Source-Free Domain Adaptation (SFDA) noise follows different distributions~\cite{yi2023source}. SFDA noisy labels are produced by adapting a pre-trained model to label the data samples. The work in~\cite{yi2023source} showed that existing noisy label learning methods relying on noise distribution assumptions are unable to address the label noise in SFDA. On the other hand, to simulate manual noisy annotations in segmentation tasks, the works~\cite{li2021superpixel,zhang2020characterizing} applied the random ratio of morphological transformation to the training data in segmentation tasks. Our work considers both SFDA noise and SFDA noise combined with morphological transformation, which are more reasonable noise patterns relevant to segmentation tasks. Moreover, we also evaluate our methods on real-world annotation noise caused by different annotators. Examples of noise types used in our work are visualized in Fig.~\ref{watch_noise}.

\section{Methodology}
In this section, we introduce our Collaborative Learning with Curriculum
            Selection (CLCS) framework in detail. We first give a brief problem statement and an overview of our method in Sec.~\ref{method_intro}, and then explain the two main modules of our method, \textit{i.e.}, the Curriculum Noisy Label Sample Selection (CNS) module and the Noise Balance Loss (NBL) module, in Sec.~\ref{method_CNS} and Sec.~\ref{method_NBL}, respectively.

\subsection{Overview}
\label{method_intro}
In the context of medical image segmentation with noisy labels, we are given a set of medical images $\mathcal{X}=\{x_i\}_{i=1}^{N}$ and the corresponding annotations $\mathcal{Y}=\{y_i\}_{i=1}^{N}$. The input image $x_i \subset \mathbb{R}^{H\times W\times 3}$ has a size of $H\times W$ with $3$ channels, while $y_i \subset \left\{ 0, 1\right\}^{H\times W\times C}$ is the one-hot ground truth segmentation mask, where $C$ indicates the number of visual classes in total. Specifically, we let $x_{i, j}$ and $y_{i, j}$ denote the value and the given label of the $j$-th pixel of the $i$-th image, respectively. Note that the annotations in the training stage are accompanied by noise, \textit{i.e.}, some of the annotations are incorrect, while the clean annotations are only available during the inference stage for validating the performance.

The overview of our proposed collaborative learning with curriculum selection (CLCS) framework is shown in Fig.~\ref{fig2}. Our CLCS framework consists of two main modules, a curriculum noisy label sample selection (CNS) module and a noise balance loss (NBL) module. In particular, our CNS module includes a boosted collaborative learning (BCL) component with a discrepancy loss, a curriculum dynamic thresholding (CDT) component, and a collaborative confidence voting (CCV) component, aiming at separating clean and noisy annotations. Then, we perform our NBL module to enable the model to make full use of both clean and noisy annotations for learning.

\subsection{Curriculum Noisy Label Sample Selection (CNS)}
\label{method_CNS}

\subsubsection{Boosted Collaborative Learning (BCL) with a Discrepancy Loss}~~~As seen from Fig.~\ref{fig2}, the core of our method is a collaborative training framework, which lays the foundation for the subsequent collaborative confidence voting component. It is based on a two-branch network while each branch independently generates predictions. Specifically, each branch consists of an encoder $\mathcal{E}^{k}$ and a decoder $\mathcal{D}^{k}$, where $k$ is the branch index. We could readily obtain the feature map extracted by each branch of the network as $f_i^k = \mathcal{E}^{k}(x_i)$ and the logits as $p_i^k = \mathcal{D}^{k}(f_i^k)$. Note that $f_i^k$ is a $D$-dimensional down-scaled feature map with a shape of $h \times w$ while $p_i^k$ is the pixel-wise logits with a shape of $H\times W\times C$. In addition, we obtain the prediction of each branch as $\hat{y}_i^k = \arg\max_c p_i^k$ and the confidence score as $\hat{p}_i^k = \max_c Softmax (p_i^k)$. 

Our two-branch network aims to generate predictions from distinct feature perspectives for a given input image. During the initial stage of training, the divergence of two branches primarily arises from random parameter initialization, as observed in Co-Teaching~\cite{han2018co} or Co-Teaching+~\cite{yu2019does}. Intuitively, this divergence enhances the network's robustness, since it provides different views, mitigating various types of errors. However, in conventional methods~\cite{han2018co,yu2019does}, the two networks tend to gradually converge as training progresses over epochs, diminishing their capacity to effectively learn and select clean data. To tackle the problem, we introduce a non-linear mapping layer $\mathcal{M}$ to the second branch of the model, creating heterogeneity in the network and generating distinct features. More importantly, we introduce a discrepancy loss designed to retain a reasonable amount of divergent predictions. In particular, we denote the features extracted by the second branch after the mapping layer as $\dot{f}_i^2 = \mathcal{M}(f_i^2)$, and propose to minimize the cosine similarity between the features from the two branches, which is termed as a discrepancy loss $\mathcal{L}_{dis}$ and defined as follows:
\begin{equation}
\mathcal{L}_{dis}=\frac{1}{N\times h\times w} \sum_{i=1}^{N} \sum_{j=1}^{h\times w}\left( 1 + \frac{f^1_{i, j} \cdot \dot{f^2_{i, j}}}{\Vert f^1_{i, j} \Vert \cdot \Vert \dot{f^2_{i, j}}\Vert} \right).
\end{equation}
Here $f_{i,j}^{k} \subset {\mathbb R}^{D\times 1}$ and $j=1, \cdots, h\times w$. The addition of ``1" is to ensure $\mathcal{L}_{dis}$ to be non-negative.

\begin{figure}[t]
\begin{center}
\centerline{\includegraphics[width=0.5\textwidth]{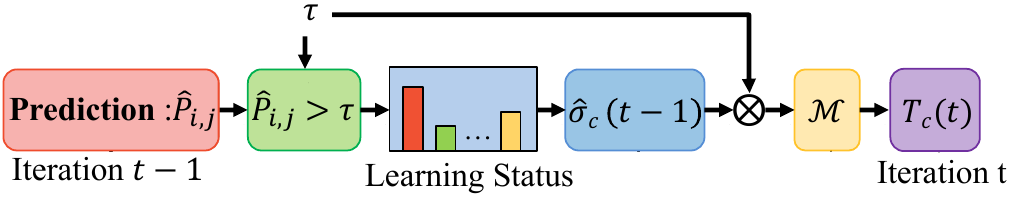}}
\end{center}
\vspace{-6mm}
\caption{\textbf{Illustration of Curriculum Dynamic Threshold (CDT). The $M$ represents a convex function.} 
} 
\label{fig-CDT}
\end{figure}

\subsubsection{Curriculum Dynamic Thresholding (CDT)}
Our BCL method is designed to generate precise predictions, emphasizing the model's capacity to learn from clean annotations. However, the inherent noise in the provided annotation set prompts us to identify confident predictions that can be utilized for further collaborative confidence voting. An ideal approach considers the predictions with a confidence score $\hat{p}_i^k$ surpassing a predetermined threshold. The threshold could be established by calculating the evaluation accuracies for each class and using the accuracies to scale a base threshold, \textit{i.e.}, $T_c(t) = A_c(t) \cdot \tau$. Unfortunately, the noisy nature of the given annotation set makes it challenging to compute an accurate $A_c(t)$, leading to an unsatisfactory outcome for confident sample selection. On the other hand, dynamically adjusting thresholds during the training process poses a significant challenge, as it substantially hinders training speed.


Facing the issues, in this part, we come up with a curriculum dynamic thresholding (CDT) strategy for confident sample selection (Fig.~\ref{fig-CDT}), which uses an alternative way to estimate the learning status, neither introducing additional inference processes nor requiring clean labels. It's crucial to note that when the threshold is set high, the learning effectiveness of a class can be inferred by examining the number of samples whose predictions falling into this class surpass the threshold. Simply put, a class with fewer samples exhibiting confidence above the threshold is perceived as having a higher noise rate, suggesting increasing learning difficulty or indicating a worse learning status~\cite{zhang2021flexmatch}. Based on this, we define the learning status of the class $c$ at time step $t$ by $\sigma_c(t)$ as:
\begin{equation}
    \sigma_c(t) = \sum_{i=1}^{N}\sum_{j=1}^{H\times W} \mathbbm{1}\left(\hat{p}_{i, j} > \tau \right) \cdot  \mathbbm{1}\left(\hat{y}_{i, j}=c\right),\label{eq3}
\end{equation}
where $\mathbbm{1} (\cdot)$ is an indicator function, returning 1 when the event occurs and 0 otherwise. In medical image segmentation tasks, lesions usually occupy a small portion of the image, which is against the substantial background component, leading to a notable class imbalance issue. The presence of noise labels compounds the challenge of label imbalance in segmentation tasks. Due to the difficulty of annotating small objects in images, there is a correlation between noise levels and object size. Therefore, the noisy label problem and the class imbalance problem become interlinked, with each reinforcing the other. This interplay manifests in their collective impact on the learning process, creating additional complexities for effective segmentation in medical image analysis.

To tackle the problem, we derive a dynamic thresholding approach that adapts the threshold to the learning status:
\begin{equation}
    T_c(t) = \hat{\sigma}_c(t) \cdot \tau, \label{eq5}
\end{equation}
where $\hat{\sigma}_c(t) = \frac{\sigma_c(t)}{\max_c \sigma_c(t)}$. Such a threshold is class-dependent. For the major class, the model often learns faster and produces more confident predictions, resulting in higher values for $\hat{\sigma}_c(t)$ and, consequently, a higher threshold $T_c(t)$. For the minor class, in turn, the lower threshold allows more model predictions to exceed the set threshold. Consequently, more minor class pixels and fewer major class pixels are picked by our dynamic threshold $T_c(t)$, reducing class imbalance.

However, during the training process, the fluctuations in $\hat{\sigma}_c(t)$ can lead to significant variations in the early phases when the model's predictions are still unstable. However, once a class is well-learned in the later stages of training, $\hat{\sigma}_c(t)$ tends to exhibit minor fluctuations. Observing this, we introduce a non-linear mapping function that shapes the thresholds into a non-linear increasing curve, which enhances the flexibility of thresholds. To achieve this, we employ a convex function $\frac{\hat{\sigma}_c(t)}{2-\hat{\sigma}_c(t)}$ to allow the thresholds to grow gradually when $\hat{\sigma}_c(t)$ is small, and become increasingly sensitive as $\hat{\sigma}_c(t)$ gets larger. Collectively, the Curriculum Dynamic Threshold (CDT) can be written as:
\begin{equation}
T_c(t) = \frac{\hat{\sigma}_c(t)}{2-\hat{\sigma}_c(t)} \cdot \tau \label{eq6}.
\end{equation}
For classes that are difficult to learn at the initial training stage, the thresholds are lowered, encouraging the selection of more training samples from these classes. As learning progresses, the thresholds for well-learned classes are raised to selectively pick up higher-quality samples. Eventually, when all classes have achieved reliable accuracies, the thresholds will stabilize.

In this way, we can obtain the masks of the selected confident predictions, denoted as:
\begin{equation}
\Delta_{i, j} = \mathbbm{1}\left( \hat{p}_{i,j} > T_{\hat{y}_{i, j}}(t) \right). \label{eq7}
\end{equation}
Note that here we omit the superscript $k$ denoting the index of the branch for simplicity.

\subsubsection{Collaborative Confidence Voting (CCV)}
Building upon our collaborative training framework, which generates predictions from diverse perspectives for each pixel, we further propose a confident voting method to identify clean label data based on the original label and the predictions of the two branches of the model. We assert that when the predictions of both branches align with the original label and exhibit high confidence, there is a strong likelihood that the label is accurate. Otherwise, we consider the corresponding annotations as potentially noisy. In particular, let $\Gamma$ indicate whether the label is clean or not, we have:
\begin{equation}
    \Gamma_{i, j} =
    \begin{cases}
    1,  & \text{$\Delta_{i, j}^1 = \Delta_{i, j}^2 = 1$ and $\hat{y}_{i, j}^1 = \hat{y}_{i, j}^2=y_{i,j}$} \\
    0, & \text{otherwise}
    \end{cases}.\label{gamma}
\end{equation}

\begin{algorithm}[t]
	\caption{CLCS Algorithm.} 
	\label{alg1} 
	\begin{algorithmic}
		\REQUIRE Training set: $\{\mathcal{X},\mathcal{Y}\}$
		\WHILE{not reach the maximum iteration} 
            \IF{ iteration $<$ warmup iteration}
            \STATE Using all pixels $\{\mathcal{X},\mathcal{Y}\}$ to calculate: $\mathcal{L}_{warmup} = \mathcal{L}_{clean} + \beta\mathcal{L}_{dis}$.\\
            \ELSE
            \FOR {c = 1 to C}
		\STATE Calculate CDT: $T_c(t) = \frac{\hat{\sigma}_c(t)}{2-\hat{\sigma}_c(t)} \cdot \tau  $. 
		\ENDFOR
            \STATE  
            Using CNS module to calculate the masks of selected confident predictions: $\Delta_{i, j}$ by Eq.~\ref{eq7}.\\
            Using CCV module to vote clean sample set: $\Gamma_{i, j}$ and noisy sample set mask: $1-\Gamma_{i, j}$by Eq.~\ref{gamma};\\
            Using NBL module to calculate $\mathcal{L}_{NBL}$ by Eq.~\ref{eq_noise} and $\mathcal{L}_{clean}$ loss by Eq.~\ref{eq_clean};\\            
            Updating the model parameters by:
            $\mathcal{L}_{total} =\mathcal{L}_{clean} + \alpha\mathcal{L}_{NBL} + \beta\mathcal{L}_{dis}.$\\
            \ENDIF
		\ENDWHILE 
        \RETURN Model parameters.
	\end{algorithmic} 
\end{algorithm}

\subsection{Noise Balance Loss (NBL)}
\label{method_NBL}
Following the distinction between clean and noisy annotations, it seems natural to focus the model's learning on the clean annotations. However, in this context, we posit that there is value in leveraging information from the noisy annotation set. The reason being, the set of noisy annotations may encompass some annotations that are correct but challenging to discern. Previous work~\cite{wang2019symmetric} has shown that cross-entropy loss is prone to make the model overfit to noisy labels on some easier classes and underlearn on some harder classes. To avoid overfitting the noisy labels, we propose a Noise Balance Loss (NBL), which can be written as:
\begin{equation}
    \begin{aligned}
    \mathcal{L}_{NBL} & = \frac{1}{2}\sum_{k=1}^2\sum_{i=1}^N \sum_{j=1}^{H\times W} [(\omega_{i, j}^k \ell_{ce}(p^k_{i, j}, y_{i, j}) \\ & + \left(1 - \omega_{i, j}^k\right) \ell_{rce}(p^k_{i, j}, y_{i, j} ))]\cdot \left( 1 - \Gamma_{i, j} \right).\label{eq_noise}
    \end{aligned}
\end{equation}
Note that our noise balance loss is a combination of the CE loss $\ell_{ce}$~\cite{zhang2018generalized} and the RCE loss $\ell_{rce}$~\cite{wang2019symmetric}, while $\ell_{ce}$ aims at achieving good convergence from the clean labels and $\ell_{rce}$ aims at mitigating the impact of noise. As demonstrated in~\cite{wang2019symmetric}, the risk minimization under $\ell_{rce}$ is noise-robust because it has a similar global minimizer under noise-free or noisy data. However, $\ell_{rce}$ exhibits weaker convergence than $\ell_{ce}$, so we combine both to benefit from their complementary nature. We design an adaptive weight to balance $\ell_{ce}$ and $\ell_{rce}$, with each weight determined by the confidence score of the corresponding prediction, denoted as $\omega_{i, j}^k = \hat{p}_{i, j}^k$. A higher confidence score signified a more likely correct prediction, leading to a higher weight assigned to $\ell_{ce}$. Conversely, lower confidence implies a higher likelihood of an incorrect prediction, leading to a higher weight to $\ell_{rce}$ to mitigate overfitting to noisy labels. This allows our approach to strike a balance between effective learning and resilience to noisy labels. Our Noise Balance Loss is exclusively utilized for optimizing the model when the annotation is identified as noisy; otherwise, the Cross-Entropy loss is employed to optimize the model. In sum, we denote the loss on the clean set as:
\begin{equation}
    \mathcal{L}_{clean} = \frac{1}{2} \sum_{k=1}^2 \sum_{i=1}^N \sum_{j=1}^{H\times W} \ell_{ce}\left(p^k_{i, j}, y_{i, j} \right) \cdot \Gamma_{i, j},\label{eq_clean}
\end{equation}
During the warm-up period, we use all data to update two networks, that is, $\Gamma_{i, j} = 1$ for all labels.

The total loss is calculated as:
\begin{equation}
    \mathcal{L}_{total} =\mathcal{L}_{clean} + \alpha \mathcal{L}_{NBL} + \beta \mathcal{L}_{dis}, \label{total}
\end{equation}
where $\alpha$ and $\beta$ are used to balance each loss term.

It should be mentioned that as demonstrated in previous work~\cite{bai2021understanding, mirzasoleiman2020coresets,chen2021robustness}, deep neural networks tend to first memorize and fit majority (clean) patterns before overfitting minority (noisy) patterns. To address this, we introduce a warmup stage at the beginning of the training process, as depicted in Alg. \ref{alg1}.

\section{Experiments}

\subsection{Implementation Details}
Our method is implemented by PyTorch and trained on a single Nvidia 2080Ti GPU. We employ DeepLabV2~\cite{chen2017deeplab} with the pre-trained encoder ResNet101 as the backbone network. The SGD optimizer is adopted. The initial learning rate is set as 1e-3. We adopt a batch size of 6 and a maximum epoch number of 200. The loss weights $\alpha$ and $\beta$ are set to 1 and 0.01, respectively, for both of the two datasets. For a fair comparison, we keep the same backbone for all baselines. For our model, we only use one branch to evaluate the model so that the model weights are at the similar level as other methods in comparison. The whole segmentation framework is trained in an end-to-end fashion. The segmentation performance is assessed by Dice and IoU scores.

\subsection{Datasets}
We validate the methods on two public datasets. The first one is the surgical instrument dataset Endovis18~\cite{allan20202018}. It consists of 2384 images annotated with three types of instrument part labels including shaft, wrist, and clasper classes. The dataset is split into 1639 training images and 596 test images following~\cite{gonzalez2020isinet}. We resize the images to 256×320 as the inputs following~\cite{guo2022joint}. The second dataset, RIGA~\cite{almazroa2017agreement}, is a benchmark for retinal cup and disc segmentation, which contains in total 750 color fundus images from three different sources, including 460 images from MESSIDOR, 195 images from BinRushed and 95 images from Magrabia. Six glaucoma experts from different organizations labeled the optic cup and disc contour masks manually for the RIGA benchmark ~\cite{almazroa2017agreement}. During model training, we selected 195 samples from BinRushed and 460 samples from MESSIDOR as the training set. The Magrabia set with 95 samples is used as the test set, which is not homologous to the training dataset. 

\subsection{Noise Patterns}
\begin{figure}[t]
\begin{center}
\centerline{\includegraphics[width=0.4\textwidth]{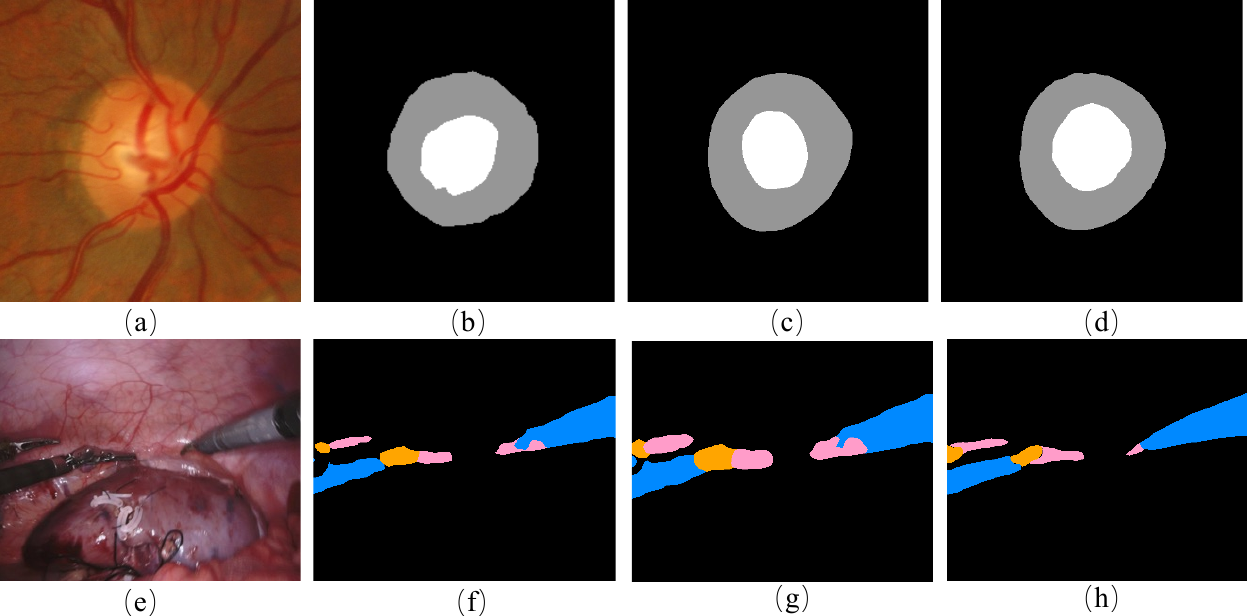}}
\end{center}
\vspace{-6mm}
\caption{\textbf{Visual comparison of different noisy labels.} Column 1: original images; Column 2: SFDA-Noise labels;
Column 3: Noise annotations from Rater 6 (fig. c) and SFDA+ED-Noise labels (fig. g); Column 4: clean segmentation labels.}
\label{watch_noise}
\end{figure}
To comprehensively verify the robustness of each method, we conduct experiments with SFDA noise, SFDA combined with morphological changes, and manual annotation noise. For the SFDA noise following~\cite{guo2022joint}, we train the source model solely on Endovis17~\cite{allan20192017} containing 1800 annotated images with domain shift to Endovis18 and generate realistic noisy labels by applying the source model directly on Endovis18. For SFDA combined with morphological changes, we introduce noises generated by randomly eroding or dilating the contours of accurate annotations following~\cite{li2021superpixel} and add them to SFDA. For the manual annotation noise, in the RIGA dataset~\cite{almazroa2017agreement}, the clean labels for the testing dataset are produced following~\cite{ji2021learning}. The noisy labels for the training dataset are from the rater 6's annotations. The above annotation noises are similar to a real-life scenario. Examples of noise patterns are shown in Fig.~\ref{watch_noise}. On the Endovis18 dataset, regarding SFDA-Noise, the noise ratios are 17.0$\%$ for Shaft, 32.1$\%$ for Wrist, and 44.3$\%$ for Clasper, while regarding SFDA+ED-Noise, the noise ratios are 20.7$\%$ for Shaft, 42.3$\%$ for Wrist, and 53.2$\%$ for Clasper. On the RIGA dataset, regarding SFDA-Noise, the noise ratio is 7.4$\%$ for Disc and 14.1$\%$ for Cup; while regarding Real-Noise, the noise ratio is 7.2$\%$ for Disc and 13.5$\%$ for Cup.

\subsection{Methods in comparison}
Three groups of methods are involved in the experimental comparison: robust loss methods, loss correction methods, and pixel-wise denoising methods. The first group, robust loss methods, designed a noise-tolerant loss term to facilitate learning of hard classes and mitigate overfitting to noisy labels. A representative method SCE~\cite{wang2019symmetric} is involved in our comparison. The second group, loss correction methods, modeled the noise transition matrix (NTM) that defines the probability of one class changing to another class. We compare our model with VolMin~\cite{li2021provably} and JACS~\cite{guo2022joint} along this line. The third group, Pixel-wise Denoising Methods, separated the dataset into the clean set and noise set, and then only trained the model on the clean set. Two representatives Co-Teaching+~\cite{yu2019does} and DCT~\cite{han2020learning} are compared. For a fair comparison, we re-run the codes released by the authors of the above works on the same training and test datasets as used by our method. The hyper-parameters of these models are reasonably tuned towards a good performance.

\begin{table*}[ht]
\centering
\caption{Comparison of segmentation results under different types of label noise on 
Endovis18~\cite{allan20202018} dataset. The baseline model is a single-branch network of our method without any label-denoising strategies.}\label{table_surgical}
\scalebox{0.75}{
\begin{tabular}{c|l|cc|cc|cc|cc}
\hline
\multirow{2}{*}{Noise Type}
& \multirow{2}{*}{Method}
&\multicolumn{2}{c|}{Shaft}&\multicolumn{2}{c|}{Wrist}&\multicolumn{2}{c}{Clasper}&\multicolumn{2}{c}{Average}\\\cline{3-10} 
&  & Dice(\%)& IoU(\%)& Dice(\%)& IoU(\%)& Dice(\%)& IoU(\%)& Dice(\%)& mIoU(\%)\\ \hline
Clean & Upper Bound  &90.74($\pm$0.09) &84.80($\pm$ 0.11) & 71.11($\pm$0.15) & 59.40($\pm$0.13)& 70.53($\pm$ 0.18) &56.61($\pm$0.19) & 74.77($\pm$0.14) & 63.65($\pm$0.14) \\ \hline
\multirow{7}{*}{SFDA-Noise} 
& Baseline  & 81.93($\pm$0.18)&72.07($\pm$0.23)&61.01($\pm$0.27)&47.30($\pm$0.24)&47.83($\pm$ 0.29)&33.59($\pm$0.30)&63.59($\pm$0.24)&50.99($\pm$0.25)\\
& SCE~\cite{wang2019symmetric}&83.02($\pm$0.15)&73.49($\pm$0.19)&61.69($\pm$0.23)&48.08($\pm$0.25)&48.79($\pm$0.22)&34.34($\pm$0.26)&64.50($\pm$0.20)&51.97($\pm$0.22)\\
& VolMin~\cite{li2021provably}&82.40($\pm$0.18)&72.71($\pm$ 0.18)&61.61($\pm$0.25)&47.90($\pm$0.23)&48.38($\pm$0.23)&34.07($\pm$0.27)&64.13($\pm$0.22) &51.56($\pm$0.22)\\
&JACS~\cite{guo2022joint}&82.28($\pm$0.19)&72.68($\pm$0.26)&60.56($\pm$0.28)&46.62($\pm$0.21)&45.56($\pm$0.24)&31.87($\pm$0.29)&62.80($\pm$0.24)&50.39($\pm$0.24)\\
&ADELE~\cite{liu2022adaptive}&82.63($\pm$0.16)&73.04($\pm$0.18)&61.65($\pm$0.18)&47.94($\pm$0.20)&49.04($\pm$0.23)& 34.82($\pm$0.26)&64.44($\pm$0.20) &51.94($\pm$0.21)\\ 
& MTCL~\cite{xu2022anti}&82.71($\pm$0.15)&72.96($\pm$0.17)&61.32($\pm$0.18)&47.63($\pm$0.23)&48.73($\pm$0.20) &34.27($\pm$0.22)&64.25($\pm$0.19)&51.62($\pm$0.22)\\
& Co-Teaching+ ~\cite{yu2019does} &82.58($\pm$0.14)&72.83($\pm$0.16)&60.96($\pm$0.20)&47.17($\pm$0.23)&49.02($\pm$0.24)&34.50($\pm$0.26)&64.19($\pm$0.21)&51.50($\pm$0.21)\\
& DCT~\cite{han2020learning}&82.88($\pm$ 0.16)&73.39($\pm$ 0.18)&61.45($\pm$ 0.22)&47.62($\pm$ 0.25)&48.44($\pm$0.25)&34.27($\pm$0.28) &64.26($\pm$0.21)&51.76($\pm$0.23)\\
& \textbf{CLCS (Ours)}&\bf85.06($\pm$0.15)&\bf77.06($\pm$0.18)&\bf63.70($\pm$ 0.20)&\bf49.78($\pm$0.22)&\bf53.77($\pm$0.22)&\bf38.80($\pm$0.24)&\bf67.51($\pm$0.19)&\bf55.21($\pm$0.21)\\ \hline
\multirow{7}{*}{SFDA + ED-Noise} 
& Baseline& 81.56($\pm$ 0.25) &71.56($\pm$ 0.31) &60.68($\pm$0.29)&46.98($\pm$0.33)&46.80($\pm$ 0.35)&32.72($\pm$0.42)&63.01($\pm$0.30)&50.42($\pm$0.34)\\
& SCE~\cite{wang2019symmetric}&82.62($\pm$0.20)&73.27($\pm$0.24) &61.29($\pm$0.31)&47.54($\pm$0.35)&49.23($\pm$0.32)&34.84($\pm$0.35)&64.38($\pm$0.28)&51.88($\pm$0.33)\\
& VolMin~\cite{li2021provably}&81.56($\pm$0.24)&71.54($\pm$0.27)&61.41($\pm$0.33)&47.70($\pm$0.36)&47.85($\pm$0.36)&33.61($\pm$0.34)&63.61($\pm$0.31)&50.95($\pm$0.32)\\
& JACS~\cite{guo2022joint}&81.73($\pm$0.26)&71.84($\pm$0.25)&60.91($\pm$0.32)&47.28($\pm$0.37)&46.93($\pm$0.37)&32.83($\pm$0.40)&63.19($\pm$0.32)&50.65($\pm$0.33)\\
& ADELE~\cite{liu2022adaptive}&81.24($\pm$0.19)&71.56($\pm$0.23)&60.65($\pm$0.27)&47.04($\pm$0.31)&50.47($\pm$0.31)&35.86($\pm$0.33)&64.12($\pm$0.26)&51.49($\pm$0.29)\\ 
& MTCL~\cite{xu2022anti}&81.55($\pm$0.22)&72.51($\pm$0.26)&61.25($\pm$0.28)&47.49($\pm$0.29)&48.29($\pm$0.32)&34.21($\pm$0.30)&64.03($\pm$0.27)&51.74($\pm$0.29)\\
& Co-Teaching+~\cite{yu2019does}&81.07($\pm$0.23)&70.82($\pm$0.26)&61.31($\pm$0.29)&47.51($\pm$0.30)&49.24($\pm$0.30)&34.88(0.31)&63.87($\pm$0.27)&51.07($\pm$0.28)\\
& DCT~\cite{han2020learning}&82.82($\pm$0.21)&73.44($\pm$0.24)&61.51($\pm$0.26)&47.64($\pm$0.28)&47.93($\pm$0.28)&34.03($\pm$0.31)&64.09($\pm$0.25)&51.70($\pm$0.27)\\
& \textbf{CLCS (Ours)}&\bf84.73($\pm$0.20)&\bf76.51($\pm$0.23)&\bf 63.31($\pm$0.27)&\bf 49.41($\pm$0.28)&\bf 53.45($\pm$0.28)&\bf 38.46($\pm$0.30)& \bf 67.15($\pm$0.25)&\bf 54.79($\pm$0.27)\\ \hline
\end{tabular}
}
\end{table*}

\begin{table*}[ht]
\centering
\caption{Comparison of segmentation results under different types of label noise on 
Endovis18~\cite{allan20202018} dataset.(The baseline model is a single-branch network of our method without any label-denoising strategies.)}\label{table_surgical_HD}
\scalebox{0.75}{
\begin{tabular}{c|l|cc|cc|cc|cc}
\hline
\multirow{2}{*}{Noise Type}
& \multirow{2}{*}{Method}
&\multicolumn{2}{c|}{Shaft}&\multicolumn{2}{c|}{Wrist}&\multicolumn{2}{c|}{Clasper}&\multicolumn{2}{c}{Average}\\\cline{3-10} 
&  & HD95 & ASSD & HD95 & ASSD& HD95 & ASSD& HD95 & ASSD\\ \hline
Clean & Upper Bound  & 7.66($\pm$0.53)&1.92($\pm$0.08)&12.02($\pm$1.06)&3.36($\pm$0.15)&13.14($\pm$1.21)&3.02($\pm$0.28)&10.94($\pm$0.93)&2.73($\pm$0.17) \\ \hline
\multirow{7}{*}{SFDA-Noise} 
& Baseline  
&19.49($\pm$1.90)&4.70($\pm$0.28)&22.18($\pm$1.23)&5.94($\pm$0.35)&34.53($\pm$2.52)&8.68($\pm$0.76)&25.39($\pm$1.88)&6.44($\pm$0.46)\\

& SCE~\cite{wang2019symmetric}
&16.23($\pm$1.80)&4.19($\pm$0.13)&18.56($\pm$0.99)&5.32($\pm$0.25)&31.06($\pm$2.13)&7.21($\pm$0.60)&21.95($\pm$1.64)&5.35($\pm$0.32)\\

& VolMin~\cite{li2021provably}
&17.05($\pm$1.84)&4.30($\pm$0.19)&18.67($\pm$1.18)&5.34($\pm$0.30)&31.25($\pm$2.05)&7.29($\pm$0.62)&22.32($\pm$1.69)&5.64($\pm$0.37)\\

&JACS~\cite{guo2022joint}
&17.52($\pm$1.86)&4.26($\pm$0.21)&20.04($\pm$1.25)&5.61($\pm$0.28)&32.18($\pm$2.64)&7.40($\pm$0.72)&23.24($\pm$1.92)&5.76($\pm$0.40)\\

&ADELE~\cite{liu2022adaptive}
&17.27($\pm$1.78)&4.19($\pm$0.17)&18.55($\pm$1.09)&5.22($\pm$0.21)&30.38($\pm$1.83)&7.15($\pm$0.50)&22.06($\pm$1.56)&5.52($\pm$0.30)\\

& MTCL~\cite{xu2022anti}
&16.62($\pm$1.69)&4.29($\pm$0.13)&19.08($\pm$1.21)&5.55($\pm$0.32)&30.64($\pm$2.24)&7.29($\pm$0.53)&22.06($\pm$1.71)&5.71($\pm$0.33)\\

& Co-Teaching+~\cite{yu2019does} 
&17.13($\pm$1.65)&4.34($\pm$0.11)&19.02($\pm$1.01)&5.33($\pm$0.26)&30.69($\pm$1.94)&7.31($\pm$0.63)&22.27($\pm$1.54)&5.66($\pm$0.33)\\ 

& DCT~\cite{han2020learning}
&16.39($\pm$1.83)&4.40($\pm$0.18)&18.39($\pm$0.94)&5.26($\pm$0.20)&29.29($\pm$1.77)&7.01($\pm$0.42)&21.36($\pm$1.84)&5.53($\pm$0.27)\\

& \textbf{CLCS (Ours)}
&13.05($\pm$1.72)&3.05($\pm$0.15)&17.41($\pm$0.89)&4.35($\pm$0.21)&25.31($\pm$1.80)&6.02($\pm$0.36)&18.59($\pm$1.47)&4.47($\pm$0.24)\\ \hline

\multirow{7}{*}{SFDA + ED-Noise} 
& Baseline& 19.57($\pm$1.92)&4.89($\pm$0.28)&23.04($\pm$1.21)&6.15($\pm$0.37)&34.87($\pm$2.64)&9.03($\pm$0.81)&25.83($\pm$1.92)&6.69($\pm$0.48)\\
& SCE~\cite{wang2019symmetric}&16.57($\pm$1.87)&4.01($\pm$0.16)&18.73($\pm$1.19)&5.52($\pm$0.32)&31.30($\pm$2.03)&7.05($\pm$0.57)&22.28($\pm$1.70)&5.53($\pm$0.35)\\
& VolMin~\cite{li2021provably}&17.12($\pm$1.85)&4.33($\pm$0.13)&19.01($\pm$1.24)&5.36($\pm$0.29)&32.87($\pm$2.22)&7.81($\pm$0.64)&23.00($\pm$1.77)&5.83($\pm$0.35)\\
& JACS~\cite{guo2022joint}&17.05($\pm$1.88)&4.13($\pm$0.17)&20.37($\pm$1.20)&5.66($\pm$0.25)&32.85($\pm$2.14)&7.53($\pm$0.60)&23.42($\pm$1.74)&5.77($\pm$0.34)\\
& ADELE~\cite{liu2022adaptive}&17.53($\pm$1.85)&4.43($\pm$0.19)&20.56($\pm$1.17)&5.71($\pm$0.30)&30.12($\pm$1.86)&6.74($\pm$0.48)&22.74($\pm$1.63)&5.63($\pm$0.32)\\ 
& MTCL~\cite{xu2022anti}&17.02($\pm$1.73)&4.09($\pm$0.14)&18.65($\pm$1.08)&5.36($\pm$0.24)&31.25($\pm$1.55)&7.03($\pm$0.53)&22.30($\pm$1.45)&5.49($\pm$0.31)\\
& Co-Teaching+~\cite{yu2019does}&17.67($\pm$1.86)&4.58($\pm$0.21)&18.64($\pm$1.13)&5.44($\pm$0.20)&30.96($\pm$1.46)&6.84($\pm$0.41)&22.42($\pm$1.48)&5.62($\pm$0.27)\\
& DCT~\cite{han2020learning}&17.05($\pm$1.81)&4.31($\pm$0.19)&18.66($\pm$1.22)&5.34($\pm$0.26)&31.23($\pm$1.62)&7.42($\pm$0.67)&22.31($\pm$1.56)&5.69($\pm$0.36)\\
& \textbf{CLCS (Ours)}&13.61($\pm$1.83)&3.48($\pm$0.16)&17.66($\pm$1.12)&4.52($\pm$0.18)&25.72($\pm$1.36)&6.31($\pm$0.38)&18.98($\pm$1.44)&4.76($\pm$0.24)\\
\hline
\end{tabular}
}
\end{table*}

\begin{figure}[ht]
    \centering
    \scalebox{0.5}{
    \includegraphics[width=1\textwidth]{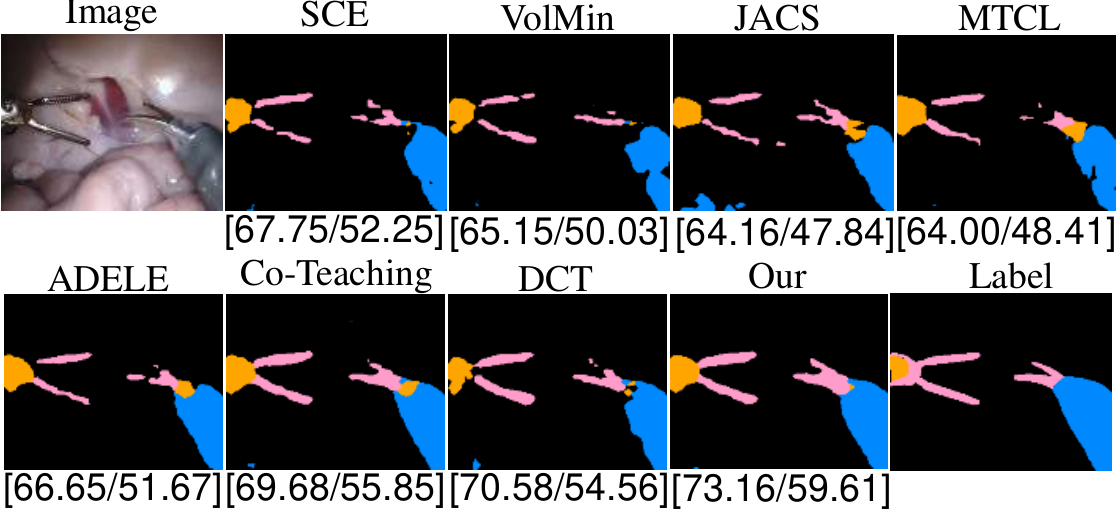}}
    \caption{Visual comparison with SFDA-Noise on Endovis18. [Dice/mIoU] are given.} 
  \label{surgical}
\end{figure}

\subsection{Performance Comparison}
In this section, we evaluate the segmentation performance of different methods on two medical image datasets with different types of noise. We also perform ablation studies to evaluate the contribution of each module of our model, i.e., CDT, CCV, discrepancy loss, and NBL.
\begin{table*}[ht]
\centering
\caption{Comparison segmentation results under different noisy label on 
RIGA~\cite{almazroa2017agreement} dataset. The baseline model is a single-branch network of our method without any label-denoising strategies.}\label{table_fundus}
\scalebox{0.8}{
\begin{tabular}{c|l|cc|cc|cc}
\hline
\multirow{2}{*}{Noise Type}
&\multirow{2}{*}{Method}
&\multicolumn{2}{c|}{Disc}&\multicolumn{2}{c|}{Cup}&\multicolumn{2}{c}{Average}\\\cline{3-8} 
&  & Dice(\%)& IoU(\%)& Dice(\%)& IoU(\%)& Dice(\%)& mIoU(\%)\\ \hline
Clean&Upper Bound&90.74($\pm$0.41)&83.25($\pm$0.25)&86.30($\pm$0.13)&77.08($\pm$0.11) & 88.52($\pm$0.27) & 80.17($\pm$0.18)\\ \hline
\multirow{7}{*}{SFDA-Noise} 
& Baseline &85.10($\pm$0.49)&74.80($\pm$0.37)&81.31($\pm$0.25)&69.95($\pm$0.28)&83.20($\pm$0.37) &72.38($\pm$0.33)\\
& SCE~\cite{wang2019symmetric}&90.00($\pm$0.43)& 82.10($\pm$0.34)&81.78($\pm$0.21)&70.84($\pm$0.23)&85.88($\pm$0.32)&76.47($\pm$0.29)\\
& VolMin~\cite{li2021provably}& 84.21($\pm$0.51)&73.57($\pm$0.42)&81.96($\pm$0.28)&70.86($\pm$0.27)&83.08($\pm$0.39)&72.22($\pm$0.35)\\
& JACS~\cite{guo2022joint}&82.95($\pm$0.50)&72.29($\pm$0.41)&80.42($\pm$0.29)&69.55($\pm$0.27)&81.68($\pm$0.39)&70.92($\pm$0.34)\\
& ADELE~\cite{liu2022adaptive}&85.55($\pm$0.45)&75.43($\pm$0.36)&82.40($\pm$0.23)&71.69($\pm$0.22)&83.97($\pm$0.34)&73.56($\pm$0.29)\\  
& MTCL~\cite{xu2022anti}&87.75($\pm$0.43)&77.75($\pm$0.33)&81.64($\pm$0.22)&71.15($\pm$0.24)&84.70($\pm$0.33)&74.45($\pm$0.29)\\
& Co-Teaching+~\cite{yu2019does}&90.73($\pm$0.44)&84.13($\pm$0.35)&78.81($\pm$0.25)&67.45($\pm$0.23)&84.77($\pm$0.34)&75.79($\pm$0.29)\\
& DCT~\cite{han2020learning}&93.69($\pm$0.42)&88.31($\pm$0.32)&77.28($\pm$0.21)&65.20($\pm$0.20)&85.49($\pm$0.32)&76.76($\pm$0.26)\\
& \textbf{CLCS (Ours)}& \bf 95.51($\pm$0.43)& \bf 91.82($\pm$0.30) &\bf 83.25($\pm$0.19) & \bf72.83($\pm$0.20) & \bf 89.38($\pm$0.31) & \bf 82.33($\pm$0.25)\\ \hline
\multirow{7}{*}{Real-Noise (Rater 6)} 
& Baseline&88.43($\pm$0.39)&79.56($\pm$0.32)&78.94($\pm$0.29)&67.33($\pm$0.30)&83.68($\pm$0.34)&73.45($\pm$0.31)\\
& SCE~\cite{wang2019symmetric}&89.80($\pm$0.35)&81.75($\pm$0.31) &80.65($\pm$0.27) &70.16($\pm$0.25) &85.22($\pm$0.31) &75.96($\pm$0.28) \\
& VolMin~\cite{li2021provably}&87.33($\pm$0.38)&77.90($\pm$0.36)&77.24($\pm$0.32)&65.42($\pm$0.30)&82.29($\pm$0.35)&71.66($\pm$0.33)\\
& JACS~\cite{guo2022joint}&88.07($\pm$0.37)&79.04($\pm$0.37)&78.57($\pm$0.30)&66.90($\pm$0.31)&83.32($\pm$0.34)&72.97($\pm$0.34)\\
&ADELE~\cite{liu2022adaptive}&88.50($\pm$0.34)&79.71($\pm$0.30)&79.34($\pm$0.27)&68.15($\pm$0.26)&83.92($\pm$0.31)&73.93($\pm$0.28)\\  
&MTCL~\cite{xu2022anti}&89.01($\pm$0.35)&80.12($\pm$0.33)&79.15($\pm$0.28)&67.98($\pm$0.27)&84.08($\pm$0.32)&74.05($\pm$0.30)\\
& Co-Teaching+~\cite{yu2019does}&94.40($\pm$0.32)&89.56($\pm$0.29)&78.47($\pm$0.26)&66.64($\pm$0.27)&86.44($\pm$0.29)&78.10($\pm$0.28)\\
& DCT~\cite{han2020learning}&95.59($\pm$0.33)&91.66($\pm$0.30)&77.31($\pm$0.25)&64.85($\pm$0.26)&86.45($\pm$0.29)&78.26($\pm$0.28)\\
& \textbf{CLCS (Ours)}& \bf 96.29($\pm$0.33)& \bf 92.91($\pm$0.32)& \bf 81.86($\pm$0.24)& \bf 70.87($\pm$0.26)& \bf 89.07($\pm$0.28)& \bf 81.89($\pm$0.29)\\ \hline
\end{tabular}
}
\end{table*}

\begin{table*}[ht]
\centering
\caption{Comparison segmentation results under different noisy label on 
RIGA~\cite{almazroa2017agreement} dataset. (The baseline model is a single-branch network of our method without any label-denoising strategies.)}\label{table_fundus_HD}
\scalebox{0.8}{
\begin{tabular}{c|l|cc|cc|cc}
\hline
\multirow{2}{*}{Noise Type}
&\multirow{2}{*}{Method}
&\multicolumn{2}{c|}{Disc}&\multicolumn{2}{c|}{Cup}&\multicolumn{2}{c}{Average}\\\cline{3-8} 
&  & HD95 & ASSD & HD95 & ASSD & HD95 & ASSD\\ \hline
Clean&Upper Bound
&10.63($\pm$0.18)&2.61($\pm$0.06)&11.03($\pm$0.22)&6.52($\pm$0.05)&10.83($\pm$0.20)&4.56($\pm$0.06)\\ \hline
\multirow{7}{*}{SFDA-Noise} 
&Baseline &12.77($\pm$0.22)&4.02($\pm$0.08)&15.23($\pm$0.25)&7.85($\pm$0.12)&14.00($\pm$0.24) &5.94($\pm$0.10)\\&SCE~\cite{wang2019symmetric}
&10.96($\pm$10.96)&3.02($\pm$0.09)&14.95($\pm$0.19)&6.57($\pm$0.09)&12.95($\pm$0.19)&4.80($\pm$0.09)\\
&VolMin~\cite{li2021provably}&12.44($\pm$0.25)&3.60($\pm$0.10)&14.53($\pm$0.16)&6.20($\pm$0.14)&13.49($\pm$0.21)&4.90($\pm$0.12)\\
&JACS~\cite{guo2022joint}&13.45($\pm$0.22)&3.76($\pm$0.07)&14.07($\pm$0.23)&7.22($\pm$0.12)&13.76($\pm$0.23)&5.49($\pm$0.10)\\
&ADELE~\cite{liu2022adaptive}&12.52($\pm$0.17)&3.64($\pm$0.08)&14.18($\pm$0.18)&5.47($\pm$0.08)&13.35($\pm$0.18)&4.56($\pm$0.08)\\  
&MTCL~\cite{xu2022anti}&11.84($\pm$0.20)&2.97($\pm$0.11)&14.25($\pm$0.16)&5.60($\pm$0.06)&13.05($\pm$0.18)&4.29($\pm$0.09)\\
& Co-Teaching+~\cite{yu2019does}&10.77($\pm$0.19)&2.29($\pm$0.08)&14.88($\pm$0.20)&5.84($\pm$0.09)&12.83($\pm$0.20)&4.07($\pm$0.09)\\
&DCT~\cite{han2020learning}&9.05($\pm$0.15)&2.49($\pm$0.08)&15.52($\pm$0.24)&7.66($\pm$0.04)&12.29($\pm$0.19)&5.08($\pm$0.06)\\
& \textbf{CLCS (Ours)}& \bf 7.81($\pm$0.15)& \bf 2.07($\pm$0.07) &\bf 12.37($\pm$0.17) & \bf3.58($\pm$0.07) & \bf 10.09($\pm$0.16) & \bf 2.83($\pm$0.07)\\ \hline
\multirow{7}{*}{Real-Noise (Rater 6)} 
&Baseline&11.4($\pm$0.18)&3.86($\pm$0.13)&14.82($\pm$0.26)&5.82($\pm$0.08)&13.11($\pm$0.22)&4.84($\pm$0.11)\\
& SCE~\cite{wang2019symmetric}&11.00($\pm$0.14)&2.75($\pm$0.05)&13.97($\pm$0.15)&4.94($\pm$0.06) &12.49($\pm$0.15) &3.85($\pm$0.06) \\
& VolMin~\cite{li2021provably}&12.04($\pm$0.21)&3.74($\pm$0.09)&15.03($\pm$0.24)&6.36($\pm$0.12)&13.54($\pm$0.23)&5.05($\pm$0.11)\\
&JACS~\cite{guo2022joint}&11.44($\pm$0.17)&3.86($\pm$0.10)&14.74($\pm$0.19)&5.69($\pm$0.10)&13.09($\pm$0.18)&4.775($\pm$0.1)\\
&ADELE~\cite{liu2022adaptive}&11.38($\pm$0.11)&3.39($\pm$0.12)&14.18($\pm$0.15)&5.05($\pm$0.09)&12.78($\pm$0.13)&4.22($\pm$0.11)\\  
&MTCL~\cite{xu2022anti}&11.56($\pm$0.16)&3.65($\pm$0.08)&14.02($\pm$0.18)&4.82($\pm$0.08)&12.79($\pm$0.17)&4.24($\pm$0.08)\\
& Co-Teaching+~\cite{yu2019does}&8.94($\pm$0.09)&2.70($\pm$0.09)&14.38($\pm$0.20)&5.69($\pm$0.11)&11.66($\pm$0.15)&4.20($\pm$0.10)\\
&DCT~\cite{han2020learning}&8.07($\pm$0.12)&2.85($\pm$0.10)&14.61($\pm$0.17)&5.77($\pm$0.06)&11.34($\pm$0.14)&4.31($\pm$0.08)\\
& \textbf{CLCS (Ours)}& \bf 7.08($\pm$0.13)& \bf 2.10($\pm$0.08)& \bf 12.84($\pm$0.16)& \bf 3.56($\pm$0.07)& \bf 9.96($\pm$0.15)& \bf 2.83($\pm$0.08)\\ \hline
\end{tabular}
}
\end{table*}

\subsubsection{Results on the Surgical Instrument Dataset}
Table~\ref{table_surgical} and Table~\ref{table_surgical_HD} present the quantitative comparison results on the Endovis18~\cite{allan20202018} training dataset with two different noise types: SFDA noise (denoted as SFDA-Noise) and SFDA noise with morphological noise (denoted as SFDA + ED-Noise). Under the typical supervised baseline setting, the baseline network (single-branch network of our method without any label-denoising strategies) achieves poor performance on both Dice and IoU on the Endovis18~\cite{allan20202018} dataset with different types of noisy labels used as ground truth labels. SCE~\cite{wang2019symmetric} using the robust loss improves the segmentation of all three parts of the surgical instrument. The loss correction methods VolMin~\cite{li2021provably} and JACS~\cite{guo2022joint} do not yield significant performance improvements, primarily due to the limitation of NTM which is too simple to learn the noise pattern effectively. Furthermore, the addition of another simple matrix as the affinity map in JACS~\cite{guo2022joint} can negatively impact the segmentation performance, especially for smaller objects such as Wrist and Clasper, as it tends to capture noisy relationships between pixels. Differently, the denoising pixel-wise label methods Co-Teaching+~\cite{han2020learning} and DCT~\cite{yu2019does} have better performance on the small objects of Wrist and Clasper than the robust loss method and the loss correction methods. However, they still lose to our model which enjoys the advantages of both the robust losses and the two-branch framework and benefits from the new collaborative confidence voting modules to produce the new SOTA results. We compare our method to the second-best performer in Table~\ref{table_surgical} in terms of Dice. For Endovis18 dataset, the p-values are 0.0071 for our method versus SCE on SFDA-Noise and 0.0014 for our method versus SCE on SFDA+ED-Noise. All p-values are less than 0.05, verifying that our method significantly outperforms the others. Visual examples are as shown in Fig~\ref{surgical}.

\subsubsection{Results on the Fundus Image Dataset}
\begin{figure}[ht]
\begin{center}
\centerline{\includegraphics[width=0.4\textwidth]{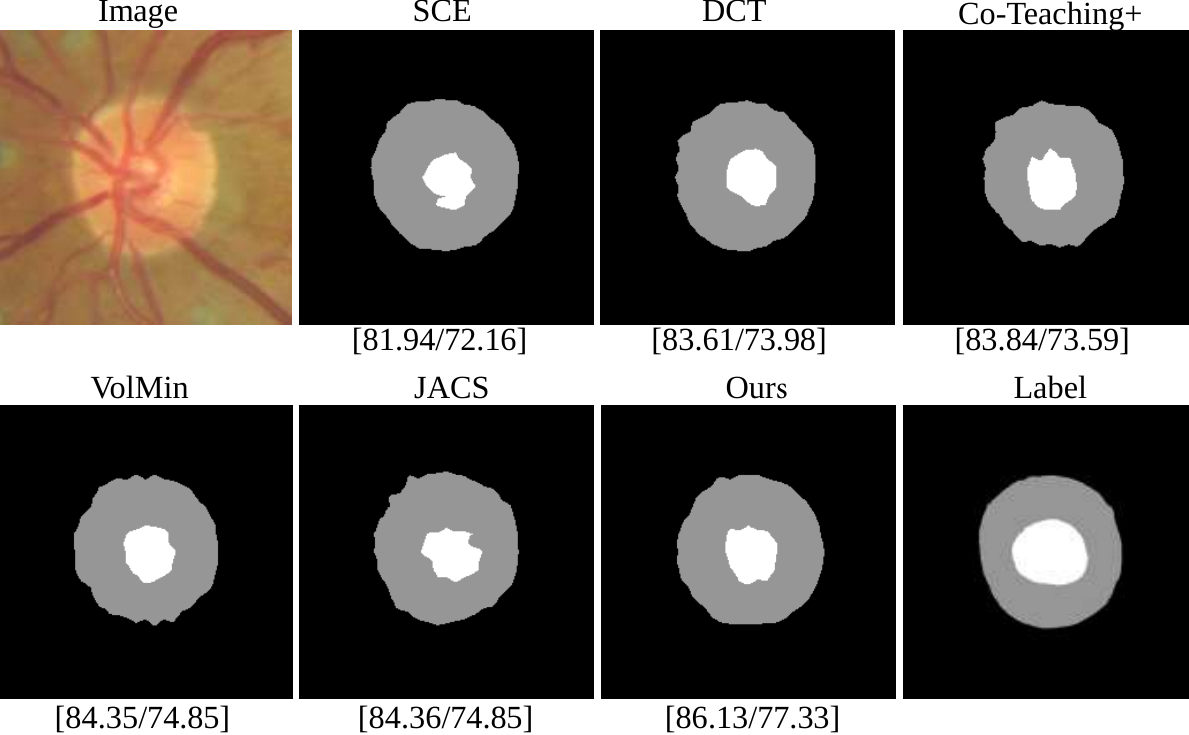}}
\end{center}
\caption{\textbf{Visual comparison of the segmentation results from different methods.} The segmentation results with Real-Noise on the training dataset. The symbol [. / .] denotes [Dice / mIoU] scores.}
\label{fundus_comparison}
\end{figure}

Table~\ref{table_fundus} and Table~\ref{table_fundus_HD} present the quantitative comparison results on the RIGA~\cite{almazroa2017agreement} dataset with two different noise types: SFDA-Noise and annotation noise by rater 6 (denoted as Real-Noise). Segmenting Disc and Cup in fundus images is inherently more challenging than segmenting surgical instruments. The former involves a deeper understanding of medical nuances, making the task more intricate with a higher prevalence of difficult pixels. The requirements for denoising methods are more demanding. Consequently, the pixel-level denoising methods Co-Teaching+~\cite{han2020learning} and DCT~\cite{yu2019does} outperform other compared methods. Again, our model is the best performer on both Dice and IoU. We compare our method to the second-best performer in Table~\ref{table_fundus} in terms of Dice. For RIGA dataset, the p-values are 0.0009 for our method versus SCE on SFDA-Noise and 0.0025 for our method versus DCT on Real-Noise. All p-values are less than 0.05, verifying that our method significantly outperforms the others.
Visual examples are as shown in Fig~\ref{fundus_comparison}.

\begin{table}[t] 
\centering
\caption{Ablation study of different initial $\tau$ on the Endovis18 and RIGA dataset with different types noise.}\label{tau_ablation}
\scalebox{0.8}{
\begin{tabular}{c|cc|cc|cc|cc}
\hline
Dataset&\multicolumn{4}{c|}{Endovis18} &\multicolumn{4}{c}{RIGA}\\
\hline
Noise Types&\multicolumn{2}{c|}{SFDA}&\multicolumn{2}{c|}{SFDA+ED}&\multicolumn{2}{c|}{SFDA}&\multicolumn{2}{c}{Real}\\
\hline 
$\tau$ & \textbf{Dice} & \textbf{mIoU} & \textbf{Dice} & \textbf{mIoU}
& \textbf{Dice} & \textbf{mIoU}
& \textbf{Dice} & \textbf{mIoU}
\\ 
\hline
0.95 &67.20 &54.79&66.86&54.50&88.70&81.82&88.80&81.49\\
0.90 &\textbf{67.51} &\textbf{55.21}&\textbf{67.15}&\textbf{54.79}&\textbf{89.38}&\textbf{82.33}&\textbf{89.07}&\textbf{81.89}\\
0.85 & 67.22&54.89&67.08&54.69&88.94&82.06&88.76&81.22\\
0.80 & 67.15&54.75&66.83&54.48&88.76&81.92&88.86&81.63\\
\hline
\end{tabular} 
}
\end{table}

\subsection{Impact of Curriculum Dynamic Threshold}
\begin{figure}[ht]
\begin{center}
\centerline{\includegraphics[width=0.4\textwidth]{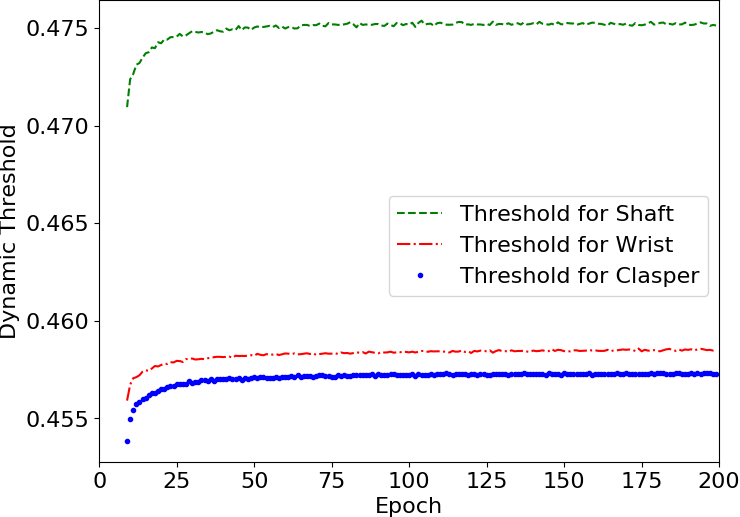}}
\end{center}
\vspace{-6mm}
\caption{\textbf{Dynamic threshold for each class.} The threshold initially increases with the number of epochs and then tends to stabilize when the model gradually stabilizes. The larger class "shaft" has a noticeably higher threshold compared to the smaller classes of "wrist" and "clasper", alleviating class imbalance in the selected clean set.
} 
\label{dynamic_threshold}
\end{figure}

If we use a fixed threshold based on the initial class distribution, i.e., with the threshold for each class being the proportion of that class's pixels to the total pixels, the Dice performance decreases from 67.51/55.21 to 65.22/52.76 on the Endovis18~\cite{allan20202018} dataset with SFDA noise. The fixed threshold cannot adapt to the model's learning status. Therefore, we employ a curriculum dynamic threshold method to adjust the threshold each epoch. Our method maintains superior performance across different $\tau$ settings in Table~\ref{tau_ablation}. We visualize how the dynamic threshold changes with the progression of training on Endovis18~\cite{allan20202018} dataset with noisy labels in Fig.~\ref{dynamic_threshold}. As shown, the threshold of each class exhibits a gradual increase during the initial 50 epochs of training, and becomes stable afterwards. In the early phases of training, the incorporation of a non-linear mapping function helps mitigate abrupt fluctuations due to the unstable model predictions. Recall that, to avoid a significant jump in the threshold during the early phases, we apply a convex function to ensure that the thresholds grow smoothly. As the model learns, it selects an increasing number of pixels as the candidates of clean subset, leading to a more stabilized distribution across class types. Beyond the 50-epoch mark, both the model's learning status and the proportion of class-specific pixels in the clean subset attain stability, resulting in the thresholds tending to plateau. The class with the largest pixel count, Shaft, naturally exhibits the highest stable threshold, aligning with our expectations. Consequently, the class imbalance is mitigated. We visualized how the number of selected pixels changes with the progression of training on the Endovis18 dataset with noisy labels in Fig.~\ref{change-of-selected-pixels1} and Fig.~\ref{change-of-selected-pixels2}. As shown, the number of selected pixels for each class increases during the training process. The increase ratios of selected pixels for each class are as follows: Shaft: 7$\%$, Wrist: 13$\%$, Clasper: 35$\%$. The increase in the number of large object pixels appears small because the total number of large object pixels is high, making the increase seem less significant. However, during training, the model learns to gradually select more difficult samples which are crucial for performance improvement. Another reason is that the noisy label ratio for large objects in segmentation tasks is lower than for small objects. The significant increase in the number of small object pixels is due to class imbalance, making small objects more challenging to learn. The model gradually learns to predict them accurately and selects the clean samples.

\begin{figure}[t]
\begin{center}
\centerline{\includegraphics[width=0.4\textwidth]{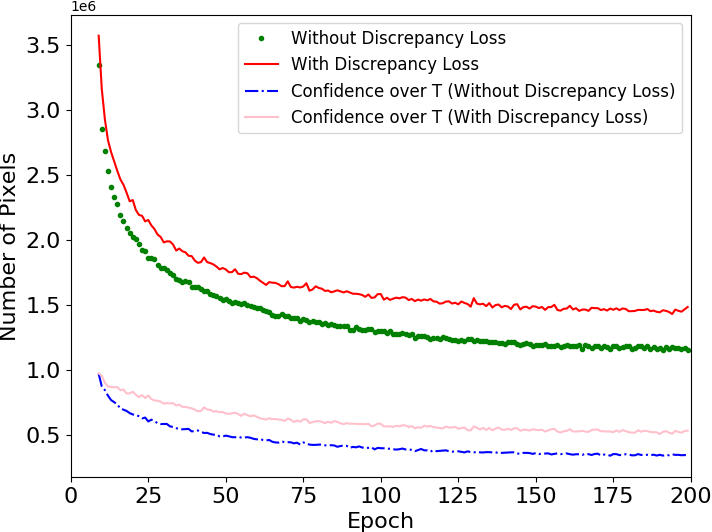}}
\end{center}
\vspace{-6mm}
\caption{\textbf{Number of pixels with different predictions by the two branches during training.} Due to the discrepancy loss, half of the total pixels have divergent predictions after 200 epochs of training.
} 
\label{two_branches}
\end{figure}
\begin{figure*}[t]
\begin{center}
\begin{minipage}{0.45\textwidth}
  \centerline{\includegraphics[width=\textwidth]{ 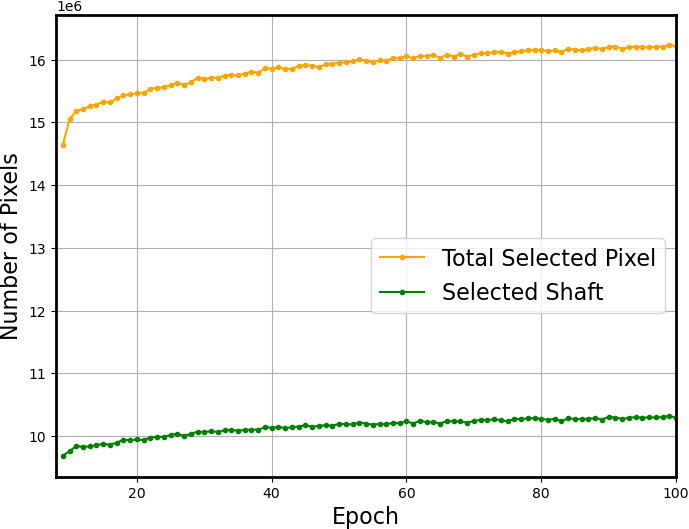}}
  \caption{\textbf{Evolution of the numbers of total selected pixels and  selected shaft pixels during training process.}}
  \label{change-of-selected-pixels1}
\end{minipage}
\hfill
\begin{minipage}{0.45\textwidth}
  \centerline{\includegraphics[width=\textwidth]{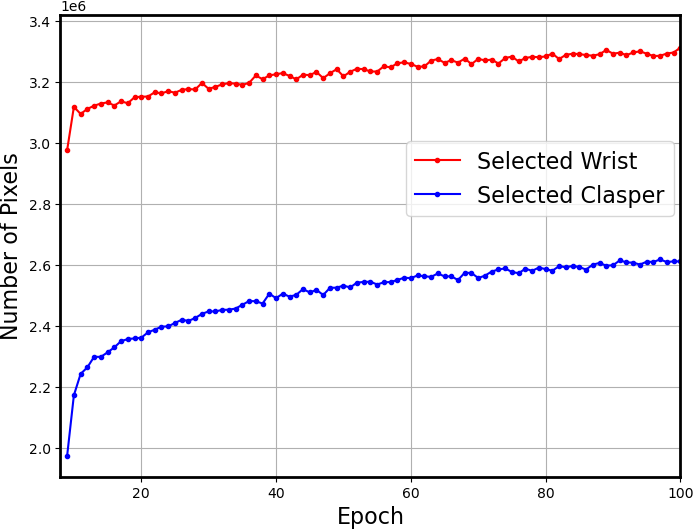}}
  \caption{\textbf{Evolution of the numbers of  selected wrist pixels and selected clasper pixels during training process.}}
  \label{change-of-selected-pixels2}
\end{minipage}
\end{center}
\end{figure*}
\begin{figure}[ht]
\begin{center}
\centerline{\includegraphics[width=0.38\textwidth]{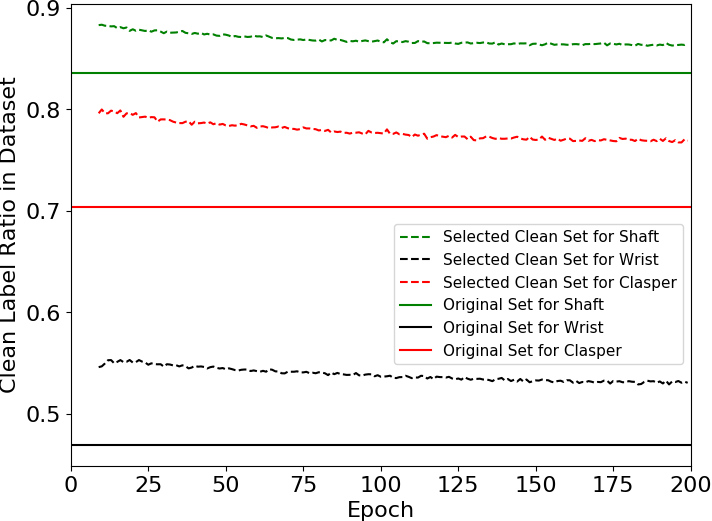}}
\end{center}
\caption{\textbf{Class-wise clean label ratios in the selected clean subset and the original dataset.} The clean label ratio is calculated as $r_{clean}(c)=\frac{\sum \mathbbm{1}(y_{clean} = y|(y=c))}{\sum \mathbbm{1}(y=c)}$, where $y_{clean}$ represents the clean labels and $y$ represents the training set labels (possibly containing noisy labels). In each class, the clean label ratio is significantly higher in the selected clean set than in the original training dataset, which enhances the segmentation performance. 
} 
\label{clean_data}
\end{figure}


\subsection{Clean Label Selection via CCV}

To verify the clean label selection through our CCV module, we plot the clean ratio curve on the Endovis18~\cite{allan20202018} dataset with noisy labels. The clean label ratio is calculated as $r_{clean}(c)=\frac{\sum \mathbbm{1}(y_{clean} = y|(y=c))}{\sum \mathbbm{1}(y=c)}$, where $y_{clean}$ represents clean labels and $y$ represents original labels (possibly containing noisy labels). Therefore, $r_{clean}(c)$ indicates the percentage of pixels with clean labels in Class $c$ for a given set. As shown in Fig.~\ref{clean_data}, the class-wise clean label ratios of the original dataset are significantly lower than those of the selected clean subset using our CCV module. Moreover, when the CCV module is applied, the improvement in clean label ratios is more noticeable for classes with a smaller number of pixels. This not only addresses the issue of noisy labels but also alleviates class imbalance. Secondly, within the clean set, there is a slight decrease in the clean ratio for each class. This is because the model inevitably fits minor noise. This decrease stabilizes as training progresses, and the extent of the decrease is very small and falls within an acceptable margin of error. In sum, the CCV module effectively separates the dataset into a clean set and noise set while considering the class imbalanced problem.

\begin{table}[t]
\centering
\caption{Model with different combination of dicrepancy loss results on the Endovis18 dataset with SFDA.}\label{JSD_ablation}
\begin{tabular}{cc|cc}
\hline
Loss Type & Level&\textbf{Dice} & \textbf{mIoU} \\ 
\hline
JSD& Prediction &65.95&53.98\\
JSD& Feature&65.72&53.67\\
Cosine& Prediction&65.81&53.26\\
Cosine& Feature&\textbf{67.51}&\textbf{55.21}\\
\hline
\end{tabular} 
\end{table}

\subsection{Impact of the Discrepancy Loss}

We now verify the effect of our discrepancy loss in encouraging different perspectives of the two branches for the same instance. Fig.~\ref{two_branches} shows the numbers of pixels predicted differently by the two branches across the training procedure based on the Endovis18~\cite{allan20202018} dataset with noisy labels. As can be seen, by using our discrepancy loss, the model tends to generate more diverse predictions compared with that not using the discrepancy loss. The number of pixels with divergent predictions remains at half of the initial values after 200 training epochs, showing the effectiveness of our discrepancy loss in preventing fast convergence of the two branches. 

We also investigate the impact of different values of the hyperparameter $\beta$ that controls the trade-off between the discrepancy loss $L_{dis}$ and the supervised losses ($L_{clean}$ and $L_{NBL}$). As shown in Table~\ref{tab2}, the proposed discrepancy loss $\mathcal{L}_{dis}$ is not too sensitive to $\beta$. The performance of the model increases when $\beta$ increases because larger $\beta$ encourages more differences between the two branches. However, if the discrepancy loss $\mathcal{L}_{dis}$ dominates, the feature learning will be negatively impacted. We set $\beta$ to 1 since it performs overall optimally in terms of most metrics

We present our method's performance using cosine similarity or Jensen-Shannon divergence (JSD) at different levels in Table~\ref{JSD_ablation}. Here,  ``feature level" indicates adding the discrepancy loss on the features from the encoder, while ``prediction level" indicates adding the discrepancy loss on the predictions from the decoder. Table~\ref{JSD_ablation} shows that using cosine similarity at the feature level yields the best performance.

\begin{table*}[ht]
\centering
\caption{Results of surgical scene segmentation (Endovis18~\cite{allan20202018} dataset) with different $\alpha$ under SFDA noise.}\label{nbl}
\begin{tabular}{c|cc|cc|cc|cc}
\hline
$\alpha$&\multicolumn{2}{c|}{Shaft}&\multicolumn{2}{c|}{Wrist}&\multicolumn{2}{c|}{Clasper}&\multicolumn{2}{c}{Average}\\
\cline{2-9} 
& Dice(\%)& IoU(\%)& Dice(\%)& IoU(\%)& Dice(\%)& IoU(\%)& Dice(\%)& mIoU(\%)\\ \hline
0.001&83.34&74.25&62.28&48.61&47.95&33.91&64.52&52.26\\
0.01 &\bf85.06&\bf 77.06&\bf63.70&\bf49.78&\bf53.77&\bf38.80&\bf67.51&\bf55.21\\
0.1 & 84.34 & 75.71 &62.93 &49.23 &51.00 &36.63 &66.09 &53.86\\
1& 83.19 &73.99 &62.37 &48.71 &48.81 &34.55 &64.79 &52.41 \\
\hline
\end{tabular}
\end{table*}

\begin{table*}[ht]
\centering
\caption{Results of surgical scene segmentation (Endovis18~\cite{allan20202018} dataset) with different $\beta$ under SFDA noise.}\label{tab2}
\begin{tabular}{c|cc|cc|cc|cc}
\hline
$\beta$&\multicolumn{2}{c|}{Shaft}&\multicolumn{2}{c|}{Wrist}&\multicolumn{2}{c|}{Clasper}&\multicolumn{2}{c}{Average}\\
\cline{2-9} 
& Dice(\%)& IoU(\%)& Dice(\%)& IoU(\%)& Dice(\%)& IoU(\%)& Dice(\%)& mIoU(\%)\\ \hline
0.01 & 83.51 & 74.42 &62.90 &49.21 &52.25 &37.65 &66.22 &53.76\\
0.1 & 83.94 & 75.13 &63.40 &49.71 &52.51 &37.93 &66.62 &54.25\\
1&\bf85.06&\bf 77.06&\bf63.70&\bf49.78&\bf53.77&\bf38.80&\bf67.51&\bf55.21 \\
10 & 84.94 & 76.82&63.58 &49.62 &53.85 &38.06 &67.45 &54.83\\
\hline
\end{tabular}
\end{table*}

\subsection{Impact of Noise Balance Loss}
From Table~\ref{ablation}, we can find that the average Dice/mIoU increases from $66.51\%/53.95\%$ to $67.51\%/55.21\%$ by adding the NBL. As shown in our loss function (Eq.~\ref{total}), a hyperparameter $\alpha$ is utilized to control the trade-off between the NBL ($\mathcal{L}_{NBL}$) and other losses ($\mathcal{L}_{clean}$ and $\mathcal{L}_{dis}$). We investigate the impact of different values of $\alpha$ as shown in Table~\ref{nbl}. As shown, the optimal performance is achieved when $\alpha$ is set as 0.01. When $\alpha$ changes within a reasonably large range, say 100 times from 0.01 to 1, our model could still outperform the methods in comparison.

\begin{table*}[ht]
\centering
\caption{Ablation study results on the Endovis18~\cite{allan20202018} dataset with SFDA + ED-Noise.}\label{ablation}
\begin{tabular}{cccccc|cc}
\hline
Baseline & $\mathcal{L}_{dis}$ & CCV & CCV(w/o mapping) & Only $\ell_{rce}$ on Noise Set & NBL & \textbf{Dice} & \textbf{mIoU}\\ 
\hline
\checkmark & & && && 63.01&50.42\\
\checkmark & \checkmark && & &&  64.18&51.71\\
\checkmark & \checkmark &  & \checkmark& & &66.01 & 53.60  \\
\checkmark & \checkmark & \checkmark &&&& 66.51 &53.95 \\
\checkmark & \checkmark & \checkmark && \checkmark & &67.18  & 54.65 \\
\checkmark & \checkmark & \checkmark &  &&\checkmark&  \bf 67.51 & \bf 55.21 \\
\hline
\end{tabular} 
\end{table*}


\subsection{Ablation Study on Model Components}

Ablation studies are performed over the key
components of the proposed CLCS, including $\mathcal{L}_{dis}$, CCV, and NBL, as reported in Table~\ref{ablation}. The ablation studies are conducted on the Endovis18~\cite{allan20202018} dataset with SFDA + ED-Noise. As seen, when we sequentially add the proposed modules to the baseline, the model performance is gradually improved. Specifically, by integrating $\mathcal{L}_{dis}$ with collaborative learning into the baseline, the two branches produce a reasonable amount of prediction disagreements, which helps vote the class probabilities of pixels in CCV. By comparing Row 3 and Row 4, we can see that the convex mapping function helps the CDT module achieve better performance. By CCV, a clean subset of candidate pixels is selected and used exclusively to train the model. This leads to a notable improvement in both Dice and mIoU, with gains of $2.33\%$ and $2.24\%$, respectively. Moreover, incorporating NBL improves data utilization, allowing those noisy samples potentially mis-detected by CCV to contribute to model learning as well. This boosts the model performance by $1.00\%$ and $1.26\%$ in the Dice and mIoU, respectively. Additionally, we show the performance of the model when only $\ell_{rce}$ is calculated on the noisy set, which is 67.18/54.65 (Dice/mIoU). The $\ell_{rce}$ is noise-robust because it has a similar global minimizer under noise-free or noisy data. Therefore, we can see the result shows that $\ell_{rce}$ is robust to the noisy label on the selected noise set. The performance improves further when we add weighted $\ell_{ce}$ to the noise set since the selected noise set contains some useful information. This demonstrates the importance of each part of the NBL module: $\ell_{ce}$ ensures learning more information and improves data efficiency, while $\ell_{rce}$ mitigates the impact of noise. Considering the small proportion of pixels ( smaller than 7\%) with noisy labels in the dataset, the extent of performance improvements is non-trivial. When considering the cumulative impact of these components, our model attains state-of-the-art performance when compared to previous methods.

\subsection{Limitation}
While our method is robust and performs exceptionally well in most medical image segmentation tasks, it may face challenges when the annotations are excessively noisy due to highly unskilled annotators. Additionally, in rare cases where there is an unusually high amount of noise, the segmentation performance of the model may be constrained. Despite these potential limitations, our method remains highly effective and reliable for the vast majority of medical image segmentation scenarios.

\section{Conclusion}

This paper introduces CLCS, a robust framework designed for noise-robust medical image segmentation. CLCS exhibits the ability to effectively learn from complex pixel-wise noisy labels while adeptly addressing the inherent class imbalance challenges associated with medical image segmentation while dealing with noisy labels. The efficacy of CLCS is validated across various types of noisy labels, including realistic annotation noise, consistently demonstrating superior performance when compared to existing methods.
\appendices
\bibliographystyle{ieeetr}
\normalem
\bibliography{mybibliography}
\end{document}